\documentclass[conference]{IEEEtran}
\usepackage{color}
\usepackage[utf8]{inputenc} % allow utf-8 input
\usepackage[T1]{fontenc}    % use 8-bit T1 fonts
\usepackage[colorlinks,citecolor=blue,linkcolor=blue,urlcolor=blue]{hyperref}
\usepackage{url}            % simple URL typesetting
\usepackage{booktabs}       % professional-quality tables
\usepackage{amsfonts}       % blackboard math symbols
\usepackage{nicefrac}       % compact symbols for 1/2, etc.
\usepackage{microtype}      % microtypography
\usepackage{booktabs}
\usepackage[table,x11names,dvipsnames,table]{xcolor}
\usepackage{graphicx}
\usepackage{wrapfig}
\usepackage[rightcaption]{sidecap}
\sidecaptionvpos{figure}{c}
\usepackage{bbm}
\usepackage{latexsym, epsfig, amssymb, amsmath, amsthm, graphicx, mathrsfs, booktabs,mathtools}
\usepackage{enumerate}
\usepackage{multirow}
\usepackage{multicol}
\usepackage{array,booktabs}
\usepackage{enumitem}
\usepackage{epstopdf}
\usepackage{caption}
\usepackage{subcaption}
\usepackage{wrapfig}
\usepackage{adjustbox}
\usepackage{multirow}
\usepackage{multicol}
\usepackage{array,booktabs}
\usepackage{algorithm}
\usepackage{algpseudocode} 
\usepackage{enumitem}
\usepackage{epstopdf}
\usepackage{caption}
\usepackage{xspace}
\usepackage{comment}
\usepackage{multirow}
\usepackage{makecell}
\usepackage{siunitx}
\usepackage{lipsum}

\newcommand{\m}[1]{{\bf{#1}}}

\DeclarePairedDelimiterX{\norm}[1]{\lVert}{\rVert}{#1}

\definecolor{darkred}{RGB}{150,0,0}
\definecolor{darkgreen}{RGB}{0,150,0}
\definecolor{darkblue}{RGB}{0,0,200}
\hypersetup{colorlinks=true, linkcolor=darkred, citecolor=darkgreen, urlcolor=darkblue}
\let\phi\varphi

\title{Truncated  Matrix Completion - An Empirical Study\\
}

\author{\IEEEauthorblockN{Rishhabh Naik, Nisarg Trivedi, Davoud Ataee Tarzanagh, Laura Balzano} %\textsuperscript{$*$}\footnote{The first two authors have contributed equally to this work.}}
\IEEEauthorblockA{\textit{Electrical and Computer Engineering} \\
\textit{University of Michigan}\\
Ann Arbor, MI, USA \\
\{rishhabh, nisargtr, tarzanaq, girasole\}@umich.edu}
}

\begin{document}
\maketitle
%\begingroup\renewcommand\thefootnote{$*$}
%\footnotetext{Equal contribution.}
%\endgroup

\begin{abstract}
%has been an active field of research since the seminal work of Cand{\`e}s and Recht proving that one may recover missing entries of partially observed low-rank matrices by solving a convex optimization problem. 
Low-rank Matrix Completion (LRMC) describes the problem where we wish to recover missing entries of partially observed low-rank matrix. 
Most existing matrix completion work deals with sampling procedures that are independent of the underlying data values. While this assumption allows the derivation of nice theoretical guarantees, it seldom holds in real-world applications. In this paper, we consider various settings where the sampling mask is dependent on the underlying data values, motivated by applications in sensing, sequential decision-making, and recommender systems. Through a series of experiments, we study and compare the performance of various LRMC algorithms that were originally successful for data-independent sampling patterns.
\end{abstract}

%Keep data-dependent sampling
%Also, it is not-so-low rank? I mean, we want to highlight something else.(= dependency)

% Previous abstract (Feb 21, removed by Laura)
%\begin{abstract}
%LRMC has been an active field of research since the seminal work of Candes and Recht proving recovery of partially observed low-rank matrices using a convex formulation under certain assumptions. Most of existing matrix-completion work deals with sampling procedures that are independent of the underlying data values. While this assumption allows derivation of nice theoretical guarantees, it seldom holds in many real-world applications. In this paper, we consider various settings where the sampling mask is dependent on the underlying data values: for example, matrix entries within a certain range are observed. Through a series of experiments, we study and compare performance of various MC algorithms that were originally successful for uniform and data-independent sampling patterns.
%\end{abstract}

\section{Introduction}
Consider the Matrix Completion (MC) problem, where the goal is to estimate an unknown $m\times n $ matrix $\m{X}^\ast$ given only few of its entries, possibly corrupted by noise. MC has found applications in diverse fields, including recommender systems~\cite{koren2009matrix}, environmental sensing~\cite{kong2013data}, sequential decision-making~\cite{grand2021robust}, computer vision~\cite{ji2010robust}, and bioinformatics~\cite{zheng2013collaborative}, to name a few. In recommender systems (RS), for example, the matrix $\m{X}^\ast$ corresponds to ratings of items (columns) by users (rows). MC in this case corresponds to predicting the ratings of all users on all items based on a few observed ratings. 

For the MC problem to be well-posed, the following assumptions are used extensively: i) $\m{X}^\ast$ is inherently of low-rank $r \ll \min(m,n)$; ii)  $\m{X}^\ast$ satisfies incoherence conditions \cite{candes2008exact}; and iii) the sample of observed entries is random and independent from the matrix $\m{X}^\ast$. Let $\Omega \subseteq [m]\times [n]$ be the subset of observed indices, and $\m{X}$ the matrix with observed entries in $\Omega$ and zero in its complement $\Omega^c$. Let $\|\m{A}\|_{\textnormal{F}(\Omega)}^2:=\sum_{(i,j)\in\Omega}a_{ij}^2$ for any matrix $\m{A}=(a_{ij})$. Then, low-rank MC (LRMC) problem is 
\begin{equation}\label{eq:LRMC}
\min_\m{Z}  \|\m{Z}-\m{X}\|_{\textnormal{F}(\Omega)}~~\mbox{subject to}~~{\mbox{rank}(\m{Z})\leq r }.
\end{equation}

In many real datasets, the low-rank assumption is at least approximately realistic. Further, low-rank approximations often yield matrices that generalize well to the unobserved entries. Incoherence is another sensible assumption; it essentially requires that key information for a good low-rank approximation can be captured with only a small sample of the matrix entries. The assumption on the independence of the sampling mask $\Omega$, however, does not often hold in real world applications \cite{agarwal2021causal}. For instance, in a movie-ratings matrix for RS, users are more likely to watch/rate movies that they will like. The probability of observing ratings for those movies would be higher than the other ones, thus violating independence from $\m{X}^\ast$. For another example, consider a collection of environmental sensors that monitor chemical levels in water. Those sensors are typically most accurate in a certain calibrated range, and outside that range the sensors will return a truncated value \cite{ni2009sensor}. These data can be treated as missing, and so in this case the missing entries are deterministically dependent on the values of $\m{X}^\ast$. %\davoud{ We give the mask formulation corresponding to these applications in Section~\ref{sec:main:result}.}

Most methods for LRMC can be assigned to one of the two classes: One class consists of algorithms that optimize over all possible $m\times n$ matrices while encouraging low-rank~\cite{candes2008exact,ma2009fixed}, whereas the second class consists of methods that explicitly enforce the rank $r$ constraint in (\ref{eq:LRMC})~\cite{bauch2021rank,keshavan2010matrix}. Several methods in the first class replace the rank constraint by a low-rank inducing penalty~\cite{candes2008exact,ma2009fixed,toh2010accelerated}. 
Most MC methods were not designed with matrix-dependent sampling patterns in mind. The purpose of this paper is to investigate their performance in this setting. 

\textbf{Our Contributions.} In this paper, we study the MC problem  with data-dependent observations, i.e., when $\Omega$ depends on $\m{X}^\ast$. Specifically, we give numerical results with three synthetic sampling patterns motivated by real-world applications that violate the standard assumption of data-independent sampling. We show that some state-of-the-art MC algorithms consistently recover $\m{X}^\ast$ under dependent sampling while others do not.

\textbf{Notation.} Vectors and matrices are denoted by bold letters, i.e., $\m{x}$ and $\m{X}$, with their elements indexed as $x_i$ and $x_{ij}$, respectively. Operations such as $\m{x} > 0$, $\m{X} \geq  0$ etc. are applied element-wise. The singular values are assumed to be arranged in non-increasing order, i.e.,  $\sigma_1\geq\sigma_2\geq \cdots \geq \sigma_r \ge 0$. For any matrix $\m{X}$, we define $\|\m{X}\|_*:= \sum_{i} \sigma_{i}^2$, $\|\m{X}\|_{\text{F}} := \sqrt{\sum_{ij}\vert x_{ij}\vert^2}$, and $\|\m{X}\|_{\text{F}(\Omega)} = \sqrt{\sum_{(i,j)\in \Omega} x_{ij}^2}$. 
% Boaz suggests we don't need to define ||-||_F{Omega} because it's defined in the left column before Eq(1). I left it since we have space.
The sampling operator $\mathcal{P}_\Omega$  extracts the entries of a matrix according to $\Omega$, such that $\mathcal{P}_\Omega$(\m X) is a vector  with entries $x_{ij}$ for $(i, j) \in \Omega$. 

%\textbf{Outline.} Section~\ref{sec:related} discusses related work, Section~\ref{sec:main:result} gives several examples of data-dependent masks, and Section~\ref{sec:experiments} provides numerical results for state of the art algorithms on these sampling masks. Finally, Section~\ref{sec:conclusion} concludes the paper.
%
%
\section{Related Work}\label{sec:related}
Related work falls into three categories: MC with data missing completely at random (MCAR), MC with data missing at random (MAR), and MC with data missing not at random (MNAR).

\noindent\textbf{MC with data missing completely at random (MCAR).}
The classical formulation and study of the LRMC problem focuses on data-independent sampling patterns with MCAR assumption~\cite{candes2008exact, bauch2021rank,Zilber2021GNMRAP,ma2009fixed, kummerle2020escaping,jain2013low}. In particular, they assume that each entry is revealed at random, independent of everything else, with \emph{uniform} probability. Under this assumption, the theoretical analysis of  LRMC methods including finite-sample consistency for both convex~\cite{candes2008exact} and non-convex formulations \cite{keshavan2010matrix} has been provided. However, MCAR is likely unrealistic due to the presence of \textit{unobserved factors} that determine both the entries of the underlying matrix and the missingness pattern in the observed matrix~\cite{agarwal2021causal}. 

\noindent\textbf{MC with data missing at random (MAR).} In this case, entries are revealed at random, independent of the underlying matrix and conditioned on observed covariates, with \textit{non-uniform} probability~\cite{little2019statistical}. For example, in RS, covariates are demographic and social network information of users, tags and content information of items. When the probabilities are non-uniform, the MC objective is biased~\cite{schnabel2016recommendations} which makes LRMC more challenging. To overcome this issue, some recent works advocate de-biasing the objective by, for example, propensity estimation techniques \cite{ma2019missing,schnabel2016recommendations}. The work in \cite{chen2015completing} proves LRMC methods succeed with sampling dependent on the leverage of the rows/columns of the matrix.
%\cite{liang2016modeling,ma2019missing,schnabel2016recommendations,wang2018modeling,wang2018collaborative,zhu2019high}.
% \noindent\textbf{MC under missing at random.} 
% Some recent work includes matrix completion with deterministic sampling patterns and assumptions to guarantee completion:
% ~\cite{kiraly2015algebraic, negahban2012restricted, chen2015completing, pimentel2016characterization, eftekhari2018mc2,liu2019matrix}.
% Few papers \cite{meka2009matrix, srebro2010collaborative} have studied low-rank matrix completion with some specific, non-uniform sampling distribution.
\\
%N: leave one empty line here for readability.
\noindent\textbf{MC with data missing not at random (MNAR).}
In this case, the missingness pattern of the matrix can be \textit{dependent} on the underlying values in that matrix, and observing the outcome of one entry can alter the probability of observing another. To address these challenges, there has been exciting recent progress on MC methods including~\cite{bhattacharya2021matrix,sportisse2020imputation,yang2021tenips} under limited version of MNAR. In particular, they assume the probability of observing each entry is a nice function solely of latent factors. 
%\cite{bhattacharya2021matrix,sportisse2020imputation,wang2020causal,yang2021tenips}. 
%These works have shown that MNAR-type algorithms outperform classical algorithms that are designed for MCAR data. However, they still assume that observation of each entry in the matrix is independent of all other entries, and each entry has a nonzero probability of being observed. 
% \\
% \cite{boumal2015low} demonstrated empirical performance of some MC algorithms under artificial sampling that seeks to mimic the distribution of data in recommender systems. Their assumptions do not generalize well, as they have created the probabilities based on specific cases.
% Some of recent work allow for MNAR data and entries of a matrix to be deterministically missing~\cite{amjad2018robust,athey2021matrix,bai2021matrix}.
% %agarwal2020principal,agarwal2019robustness,
% They consider very restricted sparsity patterns that may not be suitable for important applications of MC. 

A related literature that shares similarities with MNAR is MC with \textit{deterministic} sampling pattern \cite{athey2021matrix,bai2021matrix,pimentel2016characterization,shapiro2018matrix}. In particular, \cite{athey2021matrix,bai2021matrix} consider very restricted sparsity patterns that are not particularly suitable for important applications of MC that arise in RS or sequential decision-making. \cite{pimentel2016characterization} gave deterministic sampling conditions for unique completability and~\cite{shapiro2018matrix} provided an algebraically verifiable sufficient condition for the local uniqueness of minimum rank MC solutions. Despite sharing some similarities with MNAR, their missing pattern still does not depend on the matrix values. Our work is closely related to \cite{agarwal2021causal} where the authors proposed a causal model with provable guarantees to analyze MC with MNAR data where the probability that an entry of the matrix is missing can \emph{depend} on the underlying values in the matrix itself and depend on which other entries are missing. Motivated in part by \cite{agarwal2021causal}, we aim to investigate the efficacy of the state-of-the-art MC algorithms for recovering $\m{X}^\ast$ under various  data-independent missing patterns.

\section{ Sampling Schemes for Empirical Evaluation of Truncated Matrix Completion }\label{sec:main:result}

In this section, we describe MC problems where matrix entries are observed with a data-dependent distribution.
\subsection{ReLU-based Sampling (ReLU-S)}\label{sec:relu sampling}

In this problem, only non-negative entries of a matrix are observed. The motivation for this mask setting is two-fold: first, understanding recoverability of a ReLU-thresholded matrix can provide insights about how deep neural networks are able to learn despite the potential loss of information through multiple ReLU layers \cite{timor2022implicit}. Second, ReLU is just a special case of more generalized thresholding. Indeed, many real-world applications involve data clipped outside of certain range, e.g., voltage clipping devices \cite{ni2009sensor}. Understanding recovery of ReLU-thresholded matrices will help us in this more generic and useful case of matrices thresholded from both sides. Given $\m{X}^\ast$, let $\m{X} \in \mathbb{R}^{m \times n}$ be such that $\m X=\textnormal{ReLU}(\m X^\ast)$. Therefore the observation mask $\Omega$ is
 \begin{equation} \label{eq:2}
    (i,j) \in \Omega \hspace{1em} \textnormal{if} \hspace{1em} x^\ast_{ij} \geq 0.
 \end{equation}
%
% then aim to complete the data matrix $\m{X} \in \mathbb{R}^{m \times n}$ where $\m X=\textnormal{ReLU}(\m X^\ast)$, i.e., observations include only the non-negative entries.

%Note here we have additional information that the unobserved entries are negative.

%This setting is denoted by  ReLU in the experimental section.\\

\subsection{Group-Specific Sampling (GS-S)}
Arguably, the most well-known application of MC is RS, ubiquitous in modern online platforms. The motivation behind GS-S setting is that it seeks to mimic the structure of the data in RS. In particular, we aim to design a mask setting that reflects the self-selection bias phenomena, where most users tend to provide ratings if they particularly liked or disliked an item. However, they are much less inclined to provide a rating or even try out an item that they are lukewarm about.

 The data matrix was generated using $\m X^*=\m U\m V^\top$ and was re-scaled to have all entries between 1 and 5. This is similar to RS where users' ratings are in $[1,5]$. In GS-S setting, the mask $\Omega$ depends on the entries of $\m{X^*}$ with probabilities
\begin{equation}\label{eqn:gss}
p_{ij}= 
\begin{cases}
     0.8,& \text{if}\   x_{ij}^* \in [1,2] \cup [4,5] \\     
    0.2,
    & \text{if} \   x_{ij}^* \in (2,4). 
    % \\
    % 0.4,
    % & \text{if} \   x_{ij} \in  \\
\end{cases}
\end{equation}
This equation defines the mask $\Omega$ with smaller probability for moderate rating or entries. Note that one can use a different set of probabilities or clustering of $\m{X}$ in~\eqref{eqn:gss} for specific applications.   

\subsection{ Mean-Centric Truncated Sampling (MCT-S)} \label{sec:mean centric sampling}
Environmental sensing is a field where low-rank models are appropriate \cite{balzano2007blind,thurston1985quantitative}, but missing data issues can be so prevalent to make it too challenging to estimate low-rank models. 
The motivation behind the MCT-S setting is that it seeks to mimic the data structure of measurements from sensor nodes. For example, assume that the sensor measurements are most accurate only within a specific calibrated range. We can attempt to recover entries outside that range.

 We aim to recover entries of matrix $\m  X^{*}$ with values on either side of the matrix mean $ (mn)^{-1}\mathop{\sum_{i=1}^{m}\sum_{j=1}^{n}} x^*_{ij}$, by only observing the entries lying within a certain range. In our numerical experiments, we find an $\alpha$ so that we observe $\approx 50\%$ of the total entries, i.e.,

%In this particular case, we seek to recover missing  entries, if we observe till a range away from the mean such that, $\approx 50\%$ of the total entries 
%Let $ \m X \overset{\textnormal{i.i.d.}}{\sim} \mathcal{N}(0,\sigma^{2})$. Then, the sampling mask can be expressed as: \begin{equation} \label{eq:4}
%(i,j) \in \Omega \hspace{1em}  \textnormal{if}  \hspace{1em}  |\m X_{ij}| \leq 0.656 \sigma.
% \end{equation}
\begin{equation} \label{eq:4}
(i,j) \in \Omega \hspace{1em}  \textnormal{if}  \hspace{1em}  | x^{*}_{ij}| \leq \alpha \implies |\Omega| \approx \frac{mn}{2}\;.
 \end{equation}
\subsection{Uniformly at Random Sampling (UAR-S)}
Uniform sampling is the most common model of missingness assumed in
the MC literature \cite{candes2008exact,keshavan2010matrix,ma2009fixed, jain2013low}. Although UAR-S is likely unrealistic outside experimental settings, this regime remains a popular abstraction in machine learning and statistics to study MC problems.
%Uniform sampling is frequently used in MC where the data points are uniformly sampled across a matrix \cite{ma2009fixed, jain2013low}. 
%Although unrealistic in actual applications, this assumption has allowed for nice theoretical guarantees for recovery of partially observed matrices.
In the UAR-S scheme, the sampling is independent of the underlying distribution of the data and does not take any information from the matrix for the same. %For all our numerical experiments, we used 50$\%$ uniform sampling for fair comparison with the data-dependent schemes listed above.
%This setting is denoted by U in the experimental section.  \\

\section{Experiments} \label{sec:experiments}

This section gives simulation results demonstrating the performance of several state-of-the-art algorithms on mask settings described in Section~\ref{sec:main:result}. The following convex and non-convex MC methods are compared through experimental results:
\\

\noindent \textbf{\texttt{CVX:}} Here, we directly solve the convex relaxation of \cite{candes2008exact}, i.e., minimizing the nuclear norm subject to the linear constraints produced by the observed matrix entries. The CVX environment is a general modelling system for solving disciplined convex problems~\cite{grant2014cvx}. We used the SDPT3 solver~\cite{toh1999sdpt3}, a primal-dual interior-point algorithm that uses the path-following paradigm to solve semi-definite programming problems.
\\
\noindent\textbf{\texttt{FPCA:}} The Fixed Point Continuation with Approximate SVD algorithm \cite{ma2009fixed} solves 
 \begin{equation} \label{eq:5}
 \underset{\m {X} \in \mathbb{R}^{m \times n} }
   \min \mu ||\m X||_*  +  \frac{1}{2} ||\m {Y- X}||_F^2
 \end{equation}
using a fixed point iterative scheme through operator splitting. Here, $\mu > 0$ is a regularization parameter. The default implementation of FPCA~\cite{ma2009fixed} has a demo with UAR-S sampling.
\\
\noindent \textbf{\texttt{NNLS:}}
Nuclear Norm Least Squares  uses an accelerated proximal gradient algorithm to solve the composite problem~\eqref{eq:5}. The default implementation of NNLS \cite{toh2010accelerated} has a demo with UAR-S sampling.\\
\noindent \textbf{\texttt{R2RILS:}}  Rank $2r$ Iterative Least Squares by \cite{bauch2021rank} is a factorization based method that estimates the row and column subspaces of $\m{X}^*$, simultaneously. More specifically, given the current estimate $(\m{U}_t, \m{V}_t)$, R2RILS first solves
\begin{equation}\label{eqn:r2rils}
       \underset{\m {A} \in \mathbb{R}^{m \times n},\  {\m B \in \mathbb{R}^{m \times n}} }{\mathrm{min}} {||\m{U}_t\m{B}^\top + \m A \m V_t^\top -\m X ||_{\textnormal{F}(\Omega)} \:},
    \end{equation}
 and then uses the minimizers to obtain the new row and column subspace estimates $(\m{U}_{t+1}, \m{V}_{t+1}) $.  In the default implementation of R2RILS, $\Omega$ was generated using a Binom$(m.n,p)$. Here, $p = \rho \cdot r(m+n-r)\ /\ (m·n)\ $ and $\rho= |\Omega|\ /\ r(m+n-r)$ is oversampling ratio.
\\
\noindent\textbf{\texttt{GNMR:}} 
Gauss-Newton Algorithm for Matrix Recovery~\cite{Zilber2021GNMRAP} uses the Gauss-Newton linearization to solve the nonconvex MC problem.  More specifically,  given the current iterate $(\m {U}_t,\m {V}_t)$,  the new iterate $(\m {U}_{t+1},\m {V}_{t+1})$ is generated by solving 
%\da{USE $     \argmin$}
\begin{equation}
\underset{\m U \in \mathbb{R}^{m \times r},\m V \in \mathbb{R}^{n \times r}}{\mathrm{min}}  \ \|\mathcal P _{\Omega}(\m {U}_t \m{V}^\top + \m{U V}_t^\top -\m{U}_t \m{V}_t^\top )-b \|^2.  
\end{equation}
Here, $b$ is the vector of observed entries. In the original implementation of GNMR~\cite{Zilber2021GNMRAP}, the mask $\Omega $ was repeatedly sampled uniformly without replacement until the condition for unique recovery of the matrix~\cite{pimentel2016characterization} was satisfied.

%We would like to mention that on'
% Our initial empirical tests of SNN (Synthetic Nearest Neighbors introduced in \cite{agarwal2021causal}) yielded NRMSE (defined below) of the order of $10^{-1}$. As this NRMSE was not at par with some of the methods listed above, we did not use it for comparison in the following experiments.
% During our initial empirical tests we observed that SNN (Synthetic Nearest Neighbors introduced in \cite{agarwal2021causal}) yielded NRMSE (defined below) of the order of $10^{-1}$. We conjecture this is because we had 50\% missing entries; the SNN algorithm seemed to work well in cases where a smaller fraction of entries were missing. Due to this poor performance, we didn't include it in the following comparisons.
During our initial empirical tests we observed that SNN (Synthetic Nearest Neighbors introduced in \cite{agarwal2021causal}) did not converge for our mask settings.
We conjecture this is because we had 50\% missing entries; the SNN algorithm seemed to work well in cases where a smaller fraction of entries were missing.

\begin{figure}[h]
	\setlength{\abovecaptionskip}{0pt}
	\setlength{\belowcaptionskip}{0pt}
	\begin{center}
\includegraphics[scale=0.35]{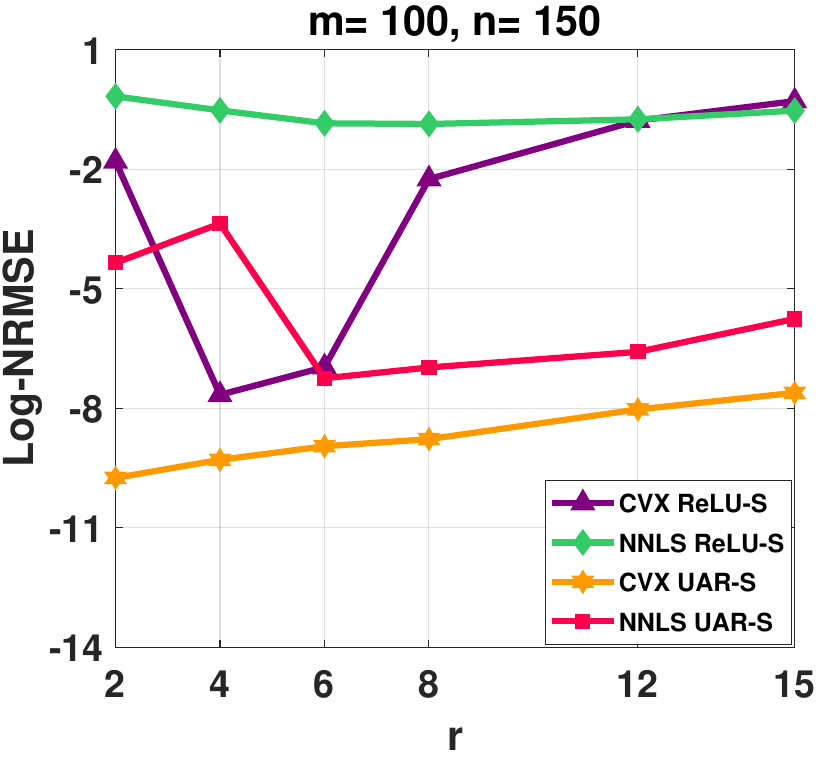}
			\caption{Performance of CVX and NNLS algorithms under ReLU-S and UAR-S schemes.}
			\label{cvxnnls}
			\end{center}
\end{figure}

Next, we discuss data generation and mask setting. For ReLU-S and MCT-S patterns, we generate a rank $r$ matrix $\m{X^{*}}$ by sampling two factors $\m U \in \mathbb{R}^{m \times r }$ and $ \m V \in \mathbb{R}^{n \times r}$ with i.i.d. Gaussian entries and set $\m{X^{*}= UV^\top}$. In the GS-S case, seeking to mimic the structure in RS, we sample $\m U \in \mathbb{R}^{m \times r }$  and $ \m V \in \mathbb{R}^{n \times r}$ with i.i.d. random entries uniformly distributed in the interval $[0,1]$ and then modify their multiplications such that each $x_{ij}$ lie between 1 and 5.
% To ensure fair comparison, we perform UAR-S the same matrices when doing each comparison (e.g. Gaussian matrices when comparing to ReLU-S/MCT-S and uniform matrices when comparing to GS-S), and we set parameters to ensure all methods observe about 50\% of the entries.

To ensure fair comparison, we set the parameters such that all methods observe about 50 \% of the entries. As discussed in Section III , the matrices for ReLU-S and MCT-S were generated using Gaussian entries while GS-S used randomly sampled uniform entries. The sampling schemes were explained in Section~\ref{sec:main:result}.

%We develop sampling masks $\Omega$ per different settings described in Section \ref{sec:main:result}. We note that in the default implementation of the above methods, demos are typically provided with a mask $\Omega $ sampled uniformly without replacement. Our experiments implemented sampling schemes of Section~\ref{sec:main:result}.

We follow the MATLAB implementations provided by the authors and observe that the default parameters provide near-optimal performance; see Table~\ref{tab:parm}.
\begin{table}[h]
\begin{center}
\begin{tabular}{ |p{.8cm}|p{5.5cm}| }
%\multicolumn{2}{|c|}{ } \\
\hline
FPCA & $\mu$ = $10^{-8}$, xtol = $10^{-6}$, $\textnormal{iter}_{\textnormal{max}}$= 500, $\tau$= 1 \\
\hline
NNLS &  $\mu$ = $10^{-8}$, xtol= $10^{-4}$, $\textnormal{iter}_{\textnormal{max}}$= 100, $\tau$= 1 \\
\hline
R2RILS & $\textnormal{t}_\textnormal{out}$= 40, $\textnormal{t}_\textnormal{in}$= 150, $\textnormal{rtol}=10^{-11}$, $\textnormal{t}_\textnormal{max}$= 40 \\
\hline
GNMR & $\textnormal{t}_\textnormal{out}$= 100, $\textnormal {t}_\textnormal{in}$= 2000,  $\textnormal {rel}_\textnormal{res} = 10^{-11}$ \\
\hline
\end{tabular}
\newline
  \caption{Parameter values for MC algorithms.}
    \label{tab:parm}
\end{center}
\end{table}

Let  $\m{X}^*$ denote true unknown matrix and $\hat{\m X}$ be the output of the above algorithms. The prediction accuracy of $\hat{\m X}$ is defined by the normalized root mean square error:
\begin{align*}
\textnormal{NRMSE}&:= \frac{\|\hat{\m X}-\m X^{*}\|_{\textnormal{F}} } {\|\m X^{*}\|_{\textnormal{F}}}~\textnormal{and}\\
~\textnormal{Log-NRMSE}&:= \log _{10} \textnormal{NRMSE}.
\end{align*}
 For each problem with $m \times n $ matrix $\m X^{*}$ with rank $r$, we solve 50 independently created MC problems for ReLU-S and MCT-S and 20 independent problems for GS-S experiments. Table~\ref{tab:parm} contains additional stopping criterion on the tolerance and the number of iterations used for convergence of the algorithms. We will consider an NRMSE of the order of $10^{-4}$ or higher as a failure case, successful otherwise. We note that in these artificial settings without added noise, recovery at machine precision is theoretically possible (NRMSE $10^{-11}$ or smaller), however we haven't used this as our success threshold, since it would not be appropriate in more realistic scenarios.
 
 %All algorithms were stopped declaring convergence at $\texttt{NRMSE} \leq 10^{-11}$. Convergence was also declared upon reaching maximum number of iterations ($\textnormal{t}_{\textnormal{max}}$) or when the relative residual ($\textnormal {rel}_{\textnormal{res}}$) between iterations was very small. We plot median $\texttt{Log-NRMSE}$ across MC experiments.
\begin{figure}[t]
	\setlength{\abovecaptionskip}{0pt}
	\setlength{\belowcaptionskip}{0pt}
	\begin{center}
		%\includegraphics[scale=0.30]{PB1_ReLU_vs_UARS_new.pdf}
		%\hspace{.1cm}
		\includegraphics[scale=0.30]{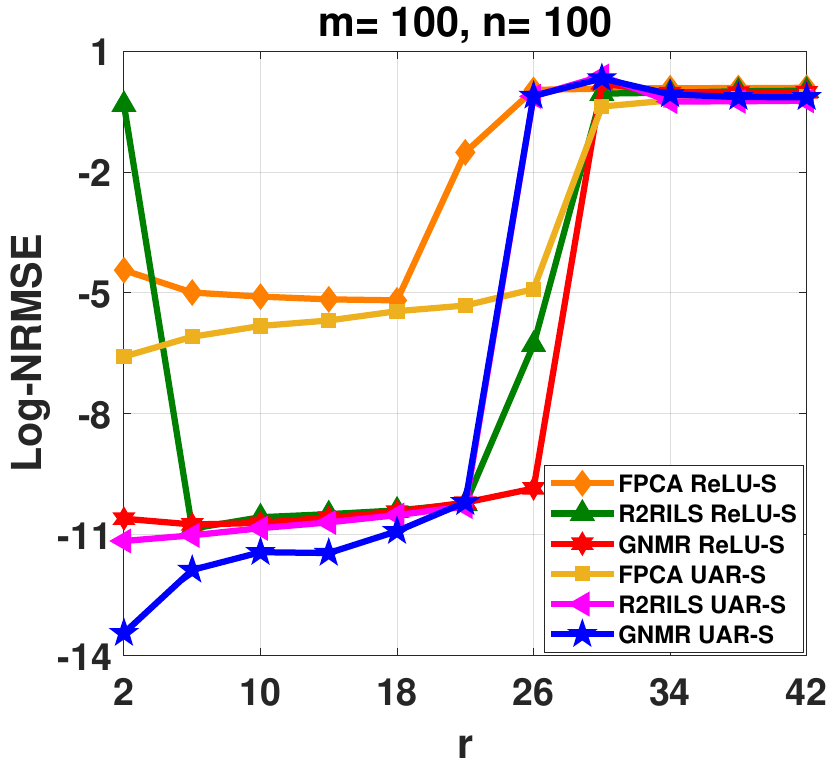}
		\hspace{.1cm}
		\includegraphics[scale=0.30]{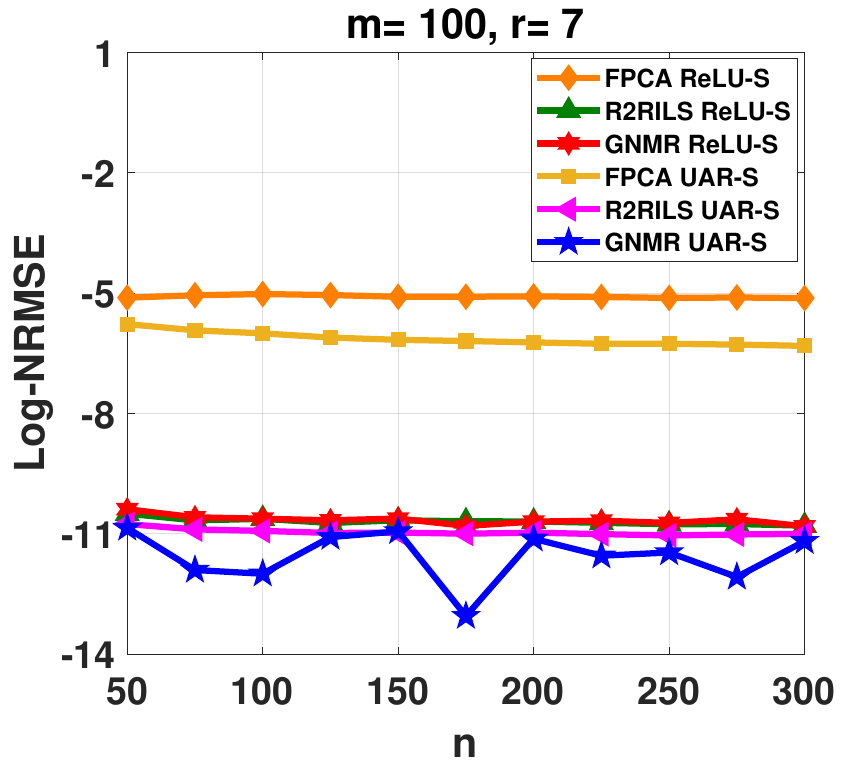}
		%\hspace{.1cm}
		%\includegraphics[scale=0.30]{PB4_ReLU_vs_UARS_new.pdf}
		%\hspace{.1cm}
		\caption{Performance of MC algorithms under ReLU-S and UAR-S schemes.}
		\label{fig:ReLU}
	\end{center}
	\vspace{-0.2cm}
\end{figure}

\begin{figure}[t]
	\setlength{\abovecaptionskip}{0pt}
	\setlength{\belowcaptionskip}{0pt}
	\begin{center}
		%\includegraphics[scale=0.30]{PB1_MCT-S_vs_UARS_new.pdf}
		%\hspace{.1cm}
		\includegraphics[scale=0.29]{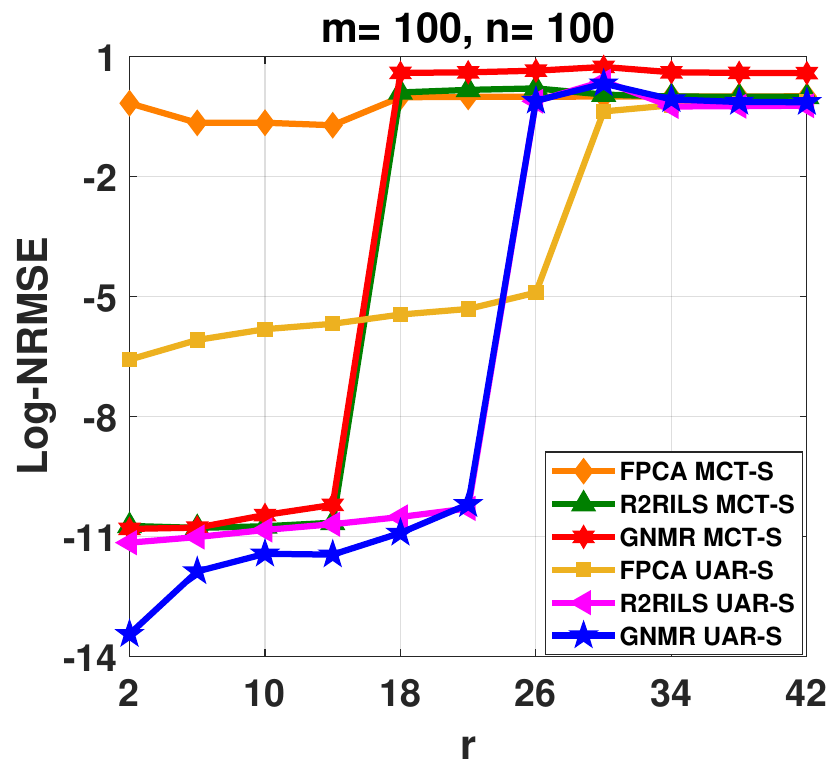}
		\hspace{.1cm}
		\includegraphics[scale=0.29]{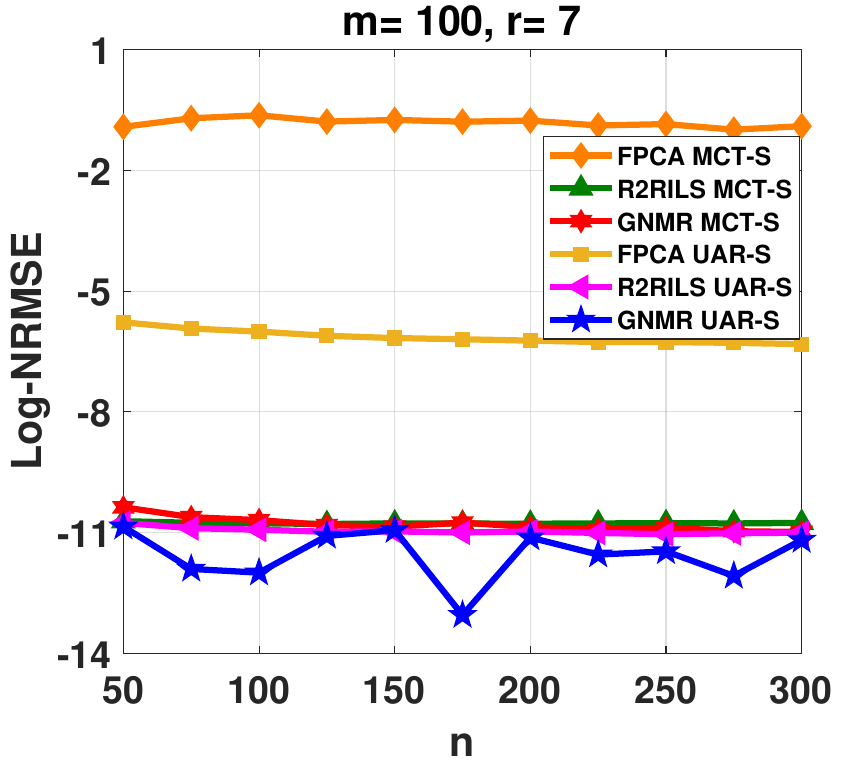}
		\hspace{.1cm}
	      %\includegraphics[scale=0.30]{PB4_MCT-S_vs_UARS_new.pdf}
		%	\hspace{.1cm}
		\caption{Performance of MC algorithms under MCT-S and UAR-S schemes.}
		\label{fig:MCTS}
	\end{center}
	\vspace{-0.2cm}
\end{figure}

%\subsection{ReLU-based Sampling (ReLU-S)
%Experiments}\label{sec:reluexpt}

%\subsubsection{CVX and NNLS}

In Figure~\ref{cvxnnls}, we compare the performance of convex optimizers {NNLS} and {CVX} for MC under the ReLU-S and UAR-S settings.
Interestingly, despite being a powerful convex solver, CVX recovers matrices only for very low-rank structures (above 2) under the ReLU-S pattern. NNLS completely fails to recover the matrices under this pattern. Our initial experiments indicated that these convex optimizers gave poor results under MCT-S and GS-S as well. Hence, we leave as future work a more extensive investigation into why these convex optimization approaches fail to consistently recover matrices when the sampling is MNAR and dependent on $\m X^\ast$.

% Interestingly, despite being a powerful convex solver, CVX recovers partially observed matrices only for a small range of ranks (above rank 2) for very low-rank structures under the ReLU-S pattern. The NNLS algorithm fails to recover the matrices even for very low-rank structures under this sampling pattern. Our initial experiments indicated that both approaches give poor results under MCT-S and GS-S. Hence, we leave as future work a more extensive investigation into why these convex optimization approaches fail to consistently recover matrices when the sampling is MNAR and dependent on $\m X^\ast$. 
%Hello Professor, we are asking why and probably also providing the reason "why these convex optimization approaches fail to consistently recover matrices when the sampling is MNAR" ...?
%Yes it would be great if we could answer that question in the future. If we can know why they fail, ideally we can create new algorithms that circumvent this issue. (Keep this in comments for our future memory :))

%No, but "we will look into why convex approaches fail for MNAR and Dependent masks"  this statement.

% Okay. Got it. Thank you. Okay.

%\subsubsection{Remaining SoTA Algorithms}
In Figure \ref{fig:ReLU}, we find that algorithms {FPCA}, {R2RILS} and {GNMR} recover low-rank matrices sampled under the ReLU-S pattern successfully when $r$ and $n$ are varied separately. Furthermore, the non-convex approaches R2RILS and GNMR recover the matrices under this pattern till  ranks similar to that in the case of UAR-S. On the other hand, FPCA, which solves the convex relaxation, begins to fail at smaller matrix rank with ReLU-S than with UAR-S.
%\subsection{ Mean-Centric Truncated Sampling (MCT-S) Experiments} \label{sec:meanexpt}

Under the MCT-S pattern in Figure~\ref{fig:MCTS}, the non-convex approaches {R2RILS} and {GNMR} with pre-specified rank $r$ recover low rank matrices successfully. FPCA, on the other hand, fails to recover matrices sampled under this pattern for even very low rank structures. The non-convex approaches although successful, have the drawback that matrices are recovered for structures of significantly lesser ranks than in the case of under UAR-S.
%\subsection{Group-Specific Sampling (GS-S) Experiments}

\begin{figure}
	\setlength{\abovecaptionskip}{0pt}
	\setlength{\belowcaptionskip}{0pt}
	\begin{center}
		\includegraphics[scale=0.3]{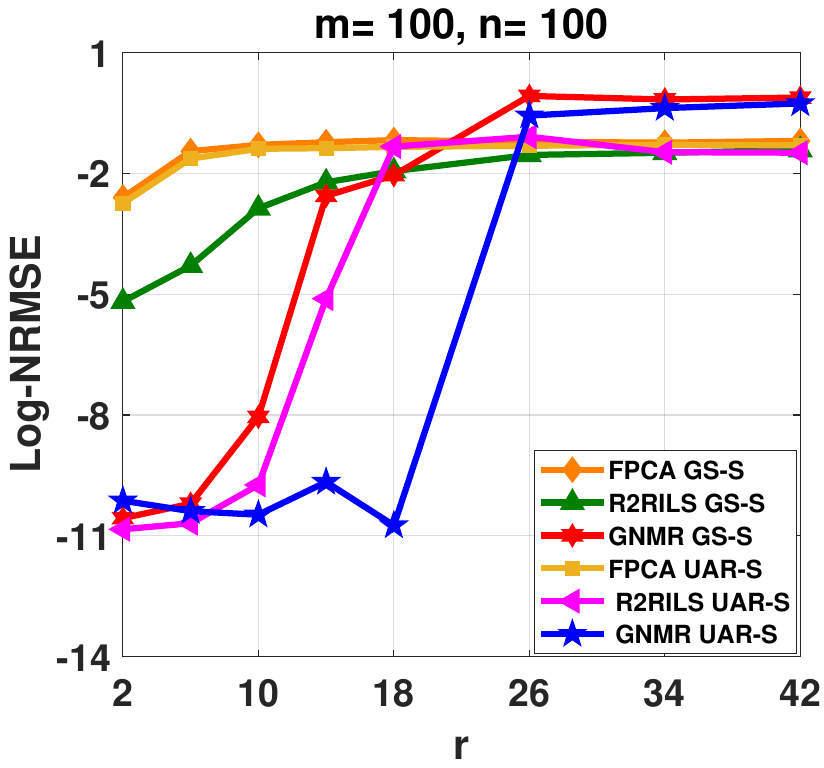}
		%\hspace{.01cm}
		\includegraphics[scale=0.3]{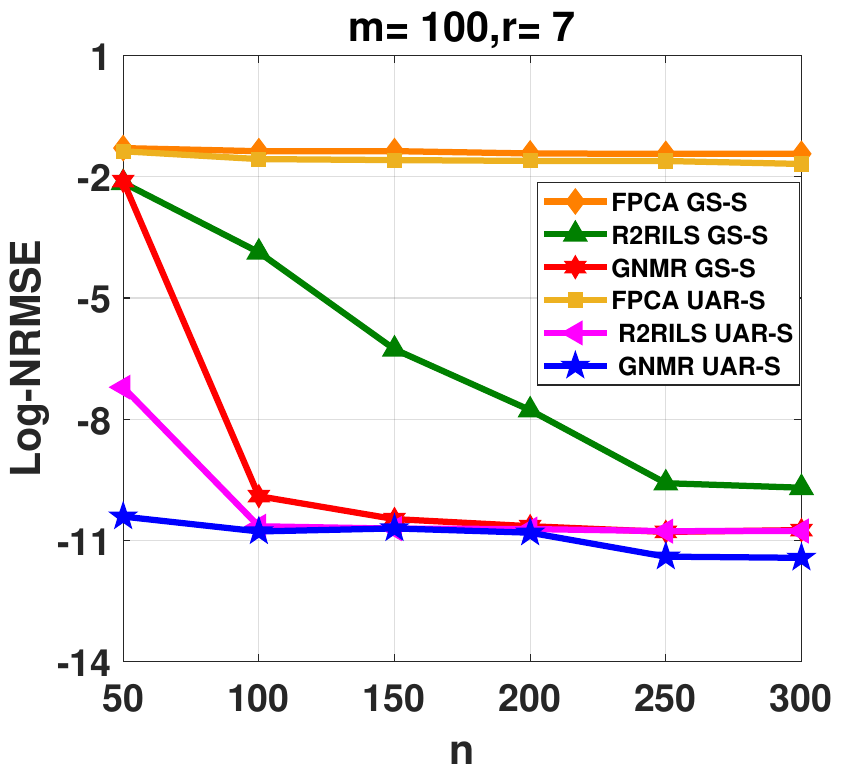}
		\caption{Performance of MC algorithms under GS-S and UAR-S schemes.}
		\label{fig:GS-S}
	\end{center}
\end{figure}

Figure \ref{fig:GS-S} shows the performance of {FPCA}, {R2RILS} and {GNMR} under GS-S pattern. One can notice that FPCA completely fails to recover the underlying matrix for both GS-S and UAR-S patterns. We believe that the failure is due to the large set of group structures in $\m{X}$ in both cases while being independent of other entries in UAR-S. This is consistent with the group-specific RS in the literature. For example, \cite{bi2017group} showed that convex MC methods can lead to a larger value of the prediction error or even failure if the data have some non-uniform group structures.  We want to highlight here that the performance on UAR-S data in Figure \ref{fig:GS-S} is different from that in earlier figures because of different data generation procedures. While comparing the performance of ReLU-S and MCT-S against UAR-S, we sample uniformly over the data matrix which is generated using low rank factorization where the matrices $\m U$ and $\m V$ have i.i.d. Gaussian entries. When comparing with GS-S, we use another UAR-S setting where the sampling is uniformly done over the data matrix generated using random i.i.d. entries and rescaled to values between 1 and 5. The entries are not normalized, unlike the UAR-S setting for the ReLU-S and MCT-S. In GS-S, we observe that FPCA fails even under the UAR-S setting, and the R2RILS performance is diminished. Similar to ReLU-S and MCT-S experiments, GNMR outperforms the rest of the algorithms under GS-S pattern. GNMR recovers the matrix even when the mask $\Omega$  is both data-dependent and MNAR for $r$ and $n$ varied separately.

We observe that across all three sampling schemes described in the paper, the non-convex approaches R2RILS and GNMR perform well across the settings, for $r$ and $n$ varied individually. GNMR performs better than the other algorithms in the GS-S setting where the structure closely resembles the nature of data in recommender systems.

%\begin{figure}[h]
		%	\includegraphics[scale=0.25]{GS-S vs UAR-S new.pdf}

		%\includegraphics[scale=0.25]{GS-S vs UAR-S new.pdf}
		%\caption{Performance of MC algorithms under GS-S and UAR-S schemes.}
%\label{figure1}
%\end{figure}

% \begin{figure*}[t]
% 	\setlength{\abovecaptionskip}{0pt}
% 	\setlength{\belowcaptionskip}{0pt}
% 	\begin{center}
% 		\includegraphics[scale=0.30]{PB1 ReLU vs UARS new.pdf}
% 		\hspace{.1cm}
% 		\includegraphics[scale=0.30]{PB2 ReLU vs UARS new.pdf}
% 		\hspace{.1cm}
% 		\includegraphics[scale=0.30]{PB3 ReLU vs UARS new.pdf}
% 		\hspace{.1cm}
% 		\includegraphics[scale=0.30]{PB4 ReLU vs UARS new.pdf}
% 		\hspace{.1cm}
% 		\caption{Performance of MC algorithms under ReLU-S and UAR-S schemes.}
% 		\label{fig:ReLU}
% 	\end{center}
% \end{figure*}

% \begin{figure*}[t]
% 	\setlength{\abovecaptionskip}{0pt}
% 	\setlength{\belowcaptionskip}{0pt}
% 	\begin{center}
% 		\includegraphics[scale=0.30]{PB1 MCT-S vs UARS new.pdf}
% 		\hspace{.1cm}
% 		\includegraphics[scale=0.30]{PB2 MCT-S vs UARS new.pdf}
% 		\hspace{.1cm}
% 		\includegraphics[scale=0.30]{PB3 MCT-S vs UARS new.pdf}
% 		\hspace{.1cm}
% 		 \includegraphics[scale=0.30]{PB4 MCT-S vs UARS new.pdf}
% 			\hspace{.1cm}
% 		\caption{Performance of MC algorithms under MCT-S and UAR-S schemes.}
% 		\label{fig:MCTS}
% 	\end{center}
% \end{figure*}

\section{Conclusion}\label{sec:conclusion} 
In this paper, we studied MC problem using three realistic sampling patterns which are data-dependent, i.e., where $\Omega$ depends on $\m{X}^\ast$. Our extensive numerical evaluations showed that MC is successful for different truncated sampling patterns, despite the standard UAR-S assumption being violated. In particular, our results indicate that non-convex SoTA algorithms {R2RILS} and {GNMR} consistently recover matrices under such realistic sampling patterns. Across all experimental settings, we find that GNMR consistently outperforms other algorithms described in this paper. Although initially designed for UAR-S, it performs well for the MC with data-dependent observations.

%with data-dependent and MNAR sampling.
%it is good, no need for "problem".
% under different it is said at the beginning "all expr." . 
% Further study is needed to derive lower bounds on the entries for recovery guarantees for finite and unique recovery of matrices subject to the ReLU and $\textbf{\textit{CM}}$ sampling patterns.
%
\section{Acknowledgements} The authors thank Avi Tachna-Fram and Anya Martin for their early contributions. This work was supported by ARO YIP award W911NF1910027 and NSF BIGDATA award IIS-1838179. 

\maketitle
%\section*{Acknowledgment}

\bibliographystyle{abbrv}%{alpha}
\bibliography{bibliography}

\end{document}